\documentclass[letterpaper, 10 pt, conference]{ieeeconf}  

\IEEEoverridecommandlockouts                              

\usepackage{graphics} 
\usepackage{epsfig} 
\usepackage{amsmath} 
\usepackage{hyperref}
\usepackage{enumerate}
\usepackage{cite} 
\usepackage{booktabs}
\usepackage{array}
\usepackage{threeparttable} 
\usepackage{xcolor,colortbl}
\definecolor{green}{RGB}{186,219,212}
\definecolor{red}{RGB}{225,180,171}
\definecolor{grey}{RGB}{229,229,229}
\definecolor{lila}{RGB}{129,15,124}
\definecolor{blue}{RGB}{4,140,212}
\definecolor{darkblue}{RGB}{4,114,178}
\newcommand{\positive}[1]{{\cellcolor{green} #1}}

\newcommand{\neutral}[1]{{\cellcolor{grey} #1}}
\usepackage{adjustbox} 
\newcommand{\asimo}{\includegraphics[scale=0.08]{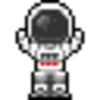}}

\newcommand{\eyesonlyc}{{\color{lila}Eyes-Only}}
\newcommand{\mirroreyesc}{{\color{blue}Mirror Eyes}}
\newcommand{\eyesonly}{{Eyes-Only}}
\newcommand{\mirroreyes}{{Mirror Eyes}}
\newcommand{\mirroreye}{{Mirror Eye}} 
\newcommand{\mycolorbox}[2]{{\setlength{\fboxsep}{1pt}\colorbox{#1}{#2}}}

\usepackage[caption=false,font=footnotesize,labelfont=sf,textfont=sf]{subfig}  

\title{\LARGE \bf
\mirroreyes: Explainable Human-Robot Interaction at a Glance}
 
\author{Matti Kr\"{u}ger\textsuperscript{\asimo}, Daniel Tanneberg\textsuperscript{\asimo}, Chao Wang\textsuperscript{\asimo}, Stephan Hasler\textsuperscript{\asimo}, and Michael Gienger\textsuperscript{\asimo}
\thanks{\asimo~Honda Research Institute Europe GmbH, Offenbach am Main, Germany.
        {\tt\small \texttt{firstname.lastname@honda-ri.de}} 
Supplementary material: \href{https://hri-eu.github.io/MirrorEyes}{hri-eu.github.io/MirrorEyes}}%
}

\usepackage{tikz}
\usepackage{eso-pic}
\usepackage{hyperref}
\newcommand\copyrighttext{%
  \footnotesize \textcopyright 2025 IEEE. Personal use of this material is permitted. Permission from IEEE must be obtained for all other uses, in any current or future media, including reprinting/republishing this material for advertising or promotional purposes, creating new collective works, for resale or redistribution to servers or lists, or reuse of any copyrighted component of this work in other works.
  Accepted to the 34th IEEE International Conference on Robot and Human Interactive Communication (RO-MAN)
}%
 
\newcommand\copyrightnotice[1][black]{%
\begin{tikzpicture}[remember picture,overlay]
\node[anchor=south,yshift=10pt,draw=#1] at (current page.south) {\parbox{\dimexpr\textwidth-\fboxsep-\fboxrule\relax}{\copyrighttext}};
\end{tikzpicture}%
}
 
\newcommand\overlaycopyrightnotice[1][black]{%
\AddToShipoutPicture*{\copyrightnotice[#1]}%
}

\begin{document}

\maketitle
\overlaycopyrightnotice 

\thispagestyle{empty}
\pagestyle{empty}

\begin{abstract}
The gaze of a person tends to reflect their interest. 
This work explores what happens when this statement is taken literally and applied to robots. 
Here we present a robot system that employs a moving robot head with a screen-based eye model that can direct the robot's gaze to points in physical space and present a reflection-like mirror image of the attended region on top of each eye. 
We conducted a user study with 33 participants, who were asked to instruct the robot to perform pick-and-place tasks, monitor the robot's task execution, and interrupt it in case of erroneous actions. 
Despite a deliberate lack of instructions about the role of the eyes and a very brief system exposure, participants felt more aware about the robot's information processing, detected erroneous actions earlier, and rated the user experience higher when eye-based mirroring was enabled compared to non-reflective eyes. 
These results suggest a beneficial and intuitive utilization of the introduced method in cooperative human-robot interaction. 
\end{abstract}

\section{INTRODUCTION}

When multiple agents work together, they can accomplish goals that would be unattainable individually. However, such cooperation typically depends on coordinated actions and mutual understanding~\cite{Bratman1992shared,Hoc2001towards,Krueger2017,Sendhoff2020,Wollstadt2022quantifying}. 
In the context of human-machine cooperation, natural language can serve as a powerful and intuitive medium for information exchange. Recent advancements in computational language processing~\cite{vaswani2017attention, brown2020language} have enabled machines to process and perform reasoning over natural language input, supporting sophisticated decision-making and action planning (e.g.,~\cite{Wang2024Lami,tanneberg2024help}). Yet, as a symbolic and context-dependent system, natural language is inherently ambiguous and susceptible to misinterpretation. 

For example, if a human requests \textit{``pass me the can on the left''}, a robot might interpret \textit{``left''} from the speaker's perspective, while the speaker actually intended the robot's left. 
Without further cues, the misunderstanding might not become evident until the action is carried out.

Multimodal communication that provides additional grounding can help to resolve some misunderstandings. For instance, in the above example, an additional glance or pointing gesture towards the object could have resolved the ambiguity. Such combinations of language and gaze or pointing gestures have been investigated previously (e.g., ~\cite{Stiefelhagen2004,Maurtua2017multimodal,Wang2020watchout,menendez2025semanticscanpathcombininggazespeech,belardinelli2025}) to enable faster and more natural communication from human to machine. Conversely, robot-generated gestures~\cite{Salem2012generation,Quintero2015,Hoffman2015design,Ishi2018} and purposeful gazing~\cite{Metta2008,Zaraki2014gazecontrol,admoni2017social,Yoshida2022,belardinelli2022intention,krueger2025virtualreflectionsdynamic2d} are being explored as communication media.

\begin{figure}
    \centering
    \includegraphics[width=1\linewidth]{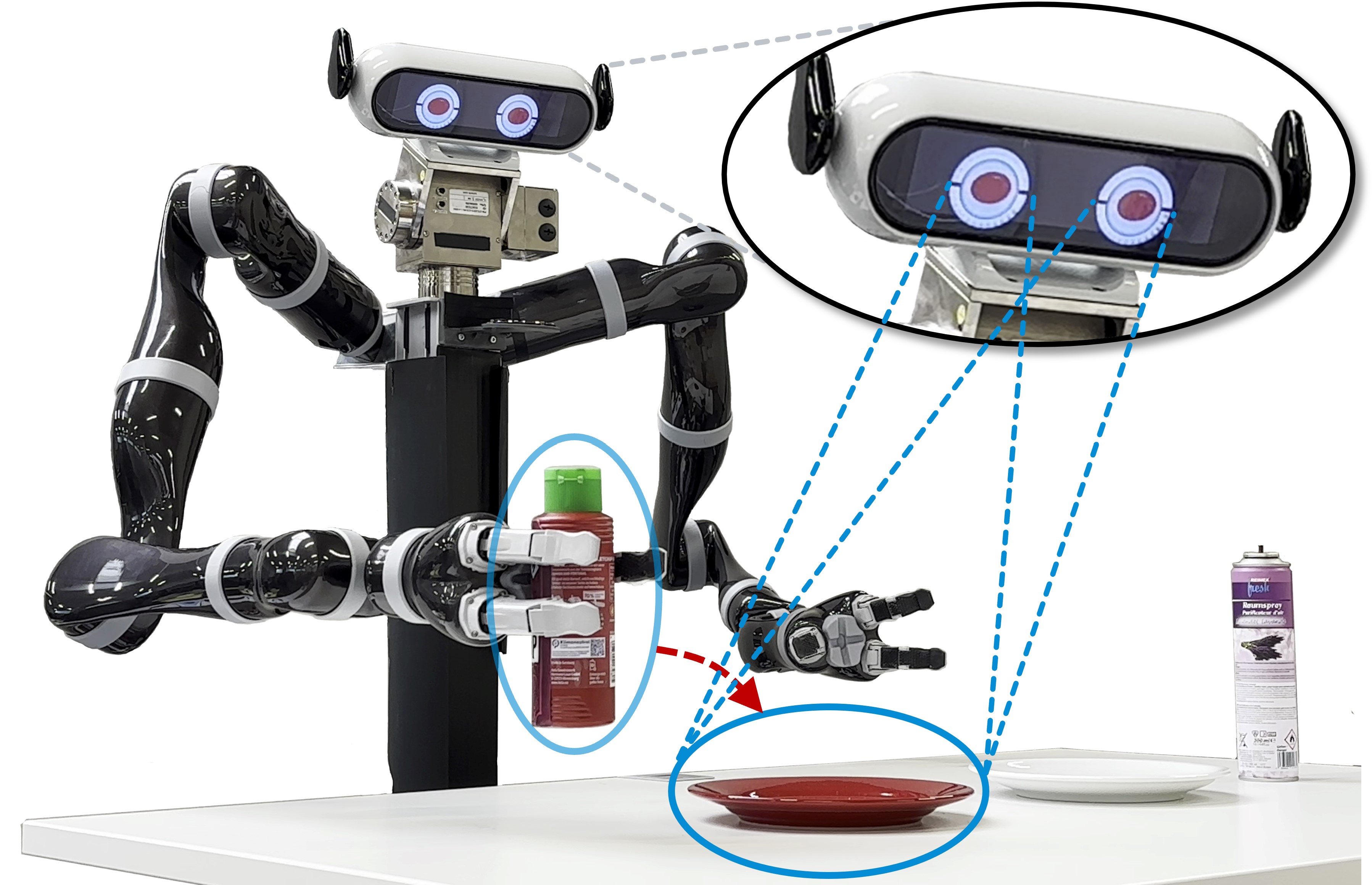}
    \caption{Robot platform with the proposed \mirroreye-enhanced head. Here the robot's task is to put the red bottle onto the red plate. In the displayed snapshot the robot has grasped the bottle and plans to put it onto the red plate next. The \mirroreyes~indicate this plan by showing the target location prior to plan execution.}
    \label{fig:robot_overview}
    \vspace{-10pt}
\end{figure}

Focusing on the task of spatial referencing with the aid of robot gaze, a critical challenge consists of conveying the gaze target vector with sufficient precision. Physical models of eye balls require robust and accurate control to generate a consistent alignment of the model's pupil with the reference target in space (e.g.,~\cite{teyssier2021eyecam}). Screen-based virtual eye models (e.g.,~\cite{Gomez2018,Yoshida2022,fang2023designing,Fang2024Roman,admoni2017social}) can reduce implementation complexity while adding expression flexibility (see, e.g., Figure~\ref{fig:eye_expressions}) but their lack of physical depth, when shown on 2D displays, introduces other challenges for creating accurately interpretable references in 3D space~\cite{krueger2025virtualreflectionsdynamic2d}.

To overcome spatial referencing limitations of screen-based eye models, Kr\"uger et al.~\cite{krueger2025virtualreflectionsdynamic2d} have introduced a reflection-like overlay for virtual eye models that presents a live-image of the current focus of attention centered on each eye's pupil. The use of this method, termed \mirroreyes, improved reference identification accuracy and user experience in a group interaction task, in which the eyes of the artificial agent were the only available reference cue.
These results offer a first glance into the efficacy of \mirroreyes~and open new avenues for further investigation. 

One such avenue is the consideration of ocular references across different object classes. 
In the study by Kr\"{u}ger et al. ~\cite{krueger2025virtualreflectionsdynamic2d}, the faces of the participants were the only attention targets that required action responses by users.
However, faces may represent a special class of attention targets, as the human brain possesses highly specialized face-processing abilities~\cite{Mccarthy1997,besson2017}, and there is evidence for a link between gaze direction and face processing~\cite{George2001} (see~\cite{krueger2025virtualreflectionsdynamic2d}).
It is unclear whether virtual mirror images of objects that lack such a dedicated neural processing architecture yield a comparable benefit. 
Secondly, the eyes were embedded in a virtual agent shown on a fixed screen that did not use any other means of spatial communication. The accuracy of screen-eye-based spatial communication could likely also be improved by embedding the eyes on mobile screens that can be physically oriented towards the attention target (e.g.,~\cite{Gomez2018}), thus partially elevating screen-based eyes to the third spatial dimension. This may lessen the need for disambiguation and call into question the necessity of additional stimuli. 

These arguments provide a basis for considering whether the previously observed benefits of \mirroreyes~can be expected to generalize to other gaze targets and more physically integrated eye models.
Therefore, we investigate the \mirroreyes~under extended conditions, where the eyes are...
\begin{enumerate}[a.]
\item implemented on a physical robot that can orient its head towards directions of interest,
\item used to attend to a wider range of object classes, and
\item used in conjunction with natural language
\end{enumerate}

We present an implementation of \mirroreyes~under these conditions and examine the following hypotheses: 

For a robot agent that generates spatial references through coordinated physical head movements and screen-based eye gaze, an addition of virtual reflections of attended objects on the eye models...
\begin{LaTeXdescription}
    \item[H1:] improves system explainability by facilitating reference disambiguation 
    \item[H2:] improves people's ability to detect situations that require constructive interference with a cooperating robot's actions 
    \item[H3:] improves the experience of object references by interaction partners 
\end{LaTeXdescription}

We further interpret the outcomes of the individual hypotheses as mutually dependent in the order presented: 
A more interpretable system (\textbf{H1}) has practical benefits for interaction (\textbf{H2}) that should raise the experience people have during interaction (\textbf{H3}). Due to the \mirroreyes' eschewal of symbolic communication, we anticipate corresponding effects to be present after brief exposure and without a need for instructions or training. 

\section{METHODOLOGY}

To investigate these hypotheses, we needed a robot system with spatial gazing capabilities, natural language interaction, basic manipulation capabilities, and the option to interfere with the system during action execution. This system is described in the following section. 

\subsection{System Description}
\label{sec:system}
We base the implementation of the robot system on our prior work \cite{joublin2024copal,tanneberg2024help} on a Large Language Model (LLM)-based robot agent that can physically and verbally interact with one or several users. The robot is depicted in Figure~\ref{fig:robot_overview}. It is a humanoid-type torso robot with two 7 degrees of freedom (DoF) arms and a pan-tilt unit carrying the expressive head. 
The head's front surface contains a 24x7~cm LCD screen that is used to display its eyes. It also holds two articulated ears to display character traits such as curiosity \cite{Wang2024Lami}. 
A 3D camera is mounted above the robot with a range large enough to see all relevant scene items and the persons located around a table in front of the robot. We use the Azure Kinect Body Tracking SDK to track the postures of interaction partners, and a combination of Segment Anything 2 \cite{ravi2024sam2}, depth information and OpenAI's multimodal gpt4o model to locate and identify objects in the scene. Spoken language and sound direction are detected with a microphone array and converted to text with Google Cloud Speech. 

The software implementation contains a set of python functions (aka ``tools'') that allow to query information about the scene, to act and manipulate the objects within its reach, and to express its internal state using cues such as speaking, pointing, and head gestures. These tools, along with a natural language text prompt describing the agent's character, constitute a run-time system that is triggered by external events (in our case detected spoken text), and responds by LLM-based reasoning about what information to gather, which tools to use, and how to reply in the given context. We found that this concept based on ``Character'' (prompt) and ``Capabilities'' (tools) in combination with advanced LLM reasoning capabilities allows natural and intuitive interactions, also comprising physical interactions. It is also flexible, making it a natural choice for the user study we present in this paper.

\subsubsection{Continuous-time Interruptibility}
We consider system interruptibility to constitute one basic form and enabler of constructive interference. Accordingly, for the present work we decided to extend the described system with the capability of being interruptible through speech. The capability uses a keyword detector that catches a set of interrupt keywords (like ``stop''). If such a keyword is detected, the system stops the interaction with the LLM, and also possibly ongoing physical interactions. On the motion planning level, we let the robot place down all objects that it currently holds in its hands. Subsequently, we stop the robot trajectories so that the robot's motion will come to a halt.

\subsubsection{Head-Eye Coordination}
\label{sec:HeadEyeCoordination}
\begin{figure}[t]
    \centering
    \includegraphics[width=0.95\linewidth]{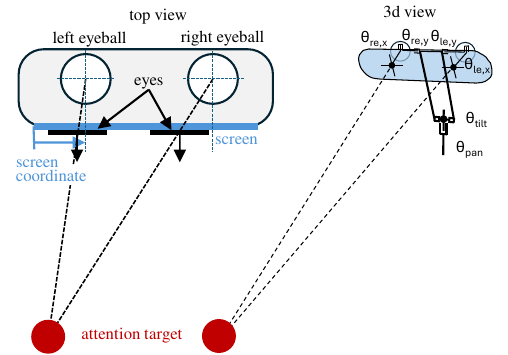}
    \caption{Kinematics of the head-eye coordination model. Left: top view, right: 3D view including neck articulation with pan and tilt joints.}
    \label{fig:eye_model}
    \vspace{-10pt}
\end{figure}

To provide a consistent real-world robot gazing behavior that allows the user to interpret the robot's spatial referencing correctly, we devised a novel kinematic gaze model with two key features: Firstly, it consistently computes the planar on-screen motion of the eyes to realistically refer to a spatial 3D scene. Secondly, it distributes the eye gaze consistently to the motion of the eyes on the screen as well as on the motor joints driving the head's motion. The model is depicted in Figure \ref{fig:eye_model}: We model two eye balls that are virtually located behind the screen. Each eye ball has two DoF, allowing its gaze vector to point at any location in the scene. We also model an attention target as a point that the system should look at. Using inverse kinematics (IK), we compute the motion of all DoFs (including the physical joints) so that the left and right eye's gaze vectors intersect the attention target. The location of the eyes on the LCD screen is then the intersection point of the corresponding eye's gaze vector with the screen surface. We use a standard differential IK with a weighted pseudo-inverse (see~\cite{nakamura1991advanced}) to compute the velocities of the DoFs: 

\begin{equation}
\dot{q} = W_q \; J^T \left(J W_q J^T \right)^{-1} e \;+\; N \; W_J \dot{q}_0
\end{equation}

Vector $q = \{\theta_{pan} \:\theta_{tilt} \: \theta_{re,x}  \:\theta_{re,y}  \: \theta_{le,x}  \: \theta_{le,y}\}^T$ comprises the physical and virtual DoFs of the model. The Jacobian $J$ contains the kinematic constraint equations that force the gaze vector to intersect the attention target, and vector $e$ is the constraint error. The diagonal weighting matrix $W_q$ is parameterized to distribute the velocities on the individual DoFs. We use low weights for the motorized neck joints, and high weights for the virtual eye-ball joints. This leads to a gaze behavior with rapidly moving eyes and an accordingly slower moving head, resulting in an impression of purposeful gazing. 
The nullspace velocity vector $q_0$ models a resting pose in the sense to bring the gaze vector into a forward-pointing direction (orthogonal to the screen). The diagonal matrix $W_J$ removes nullspace motion for the neck joints. This makes the head eventually converge to frontally face the attention target. This control concept also allows to seamlessly combine gaze with head gestures such as nodding or head shaking. To accomplish this, we simply add constraint equations for the DoFs related to the gestures in the kinematic model (typically pan and tilt DoFs). By forcing the gesture trajectory into these constraint variables, IK will adjust the gaze vector so that the gaze attention target is tracked throughout the head gesture, effectively mimicking the behavior of the vestibulo-ocular reflex (VOR)~\cite{shibata2001biomimetic}.  

\subsubsection{Eye Reflections}
\begin{figure}[t]
    \centering
    \includegraphics[width=1\linewidth]{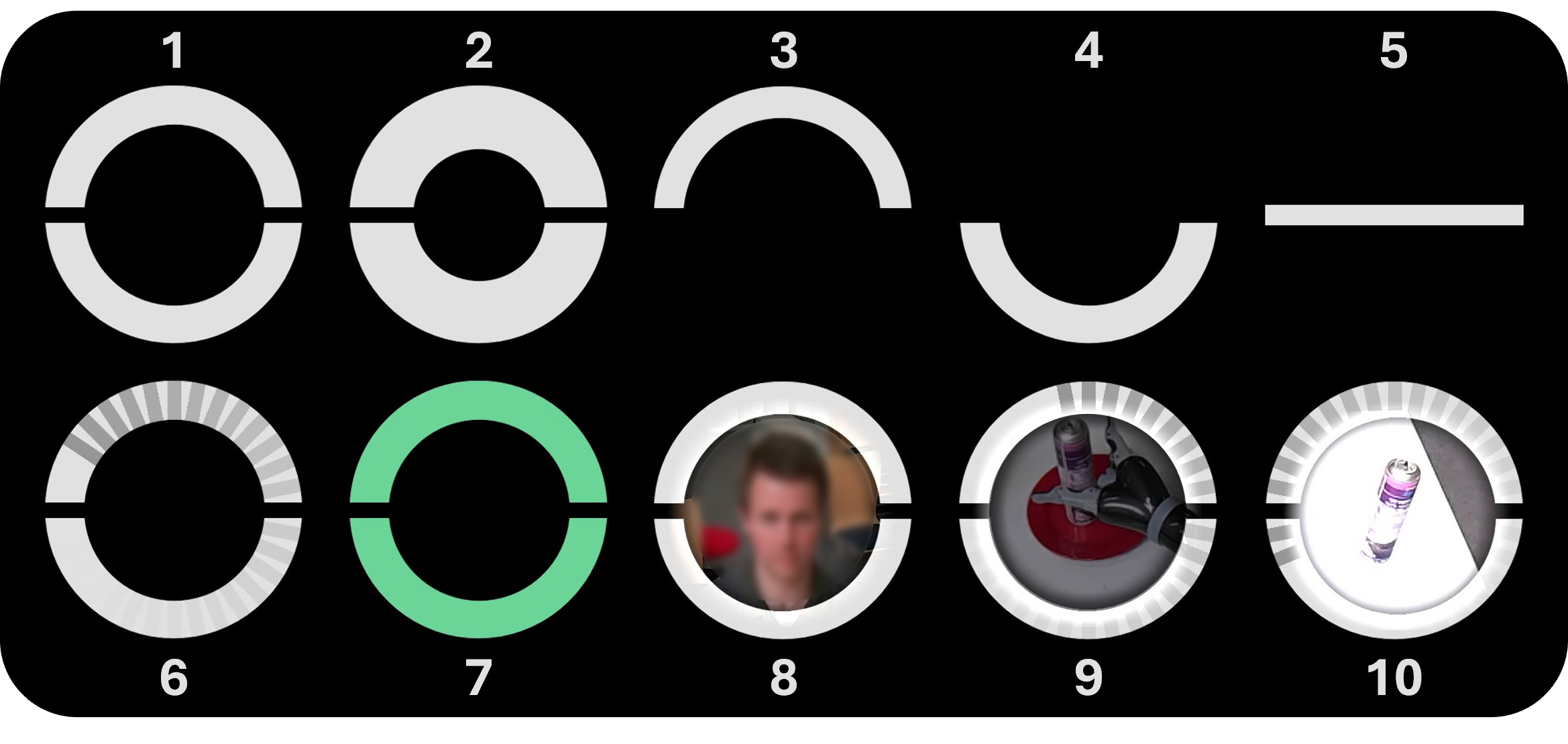}
    \caption{Selected eye-based expressions. \textit{(Top row)} Human-inspired expression primitives: 1. stylized iris and pupil, 2. reduced pupil size, 3. positive state, 4. negative state, 5. eyes closed, \textit{(Bottom row)} Superhuman augmentations: 6. processing animation, 7. color coding of robot state, 8. \mirroreyes~focused on a person (blurred), 9. \mirroreyes~focused on objects at lower reflection opacity + loading animation indicating processing, 10. \mirroreyes~focus on an object with a brief overexposure (flash) indicating first registration.}
    \label{fig:eye_expressions}
\end{figure}
To realize the \mirroreye~effect, we added a mirror module that complements the rendering of the eye projections by creating an overlay of a horizontally flipped segment of a camera feed at a chosen transparency level. This module is based on the virtual reflection module introduced by Kr\"{u}ger et al.~\cite{krueger2025virtualreflectionsdynamic2d} but requires a different method for pupil-image alignment due to the underlying eye movement control described in Section~\ref{sec:HeadEyeCoordination}. Additionally, we extended the mirror module to be able to dynamically scale camera images of attended regions to the pupil size of the eyes in the case of attending objects of varying size or distance to the camera. These changes resulted in the following method for pupil-image alignment: 

Given a point of interest $(p_x, p_y)$ on a camera image with size $c_w$ x $c_h$, the segment of its (optionally rescaled) mirror image with the top left corner $(tl_x, tl_y)$ and target size $t_w$ x $t_h$ for rendering on the pupils, is given by 

\begin{equation}
\label{eq_tl}
tl_{x|y} = o_{x|y} + \left(s \cdot t_{w|h} \cdot p_{x|y}\right) - t_{w|h}
\end{equation}

In case of rescaling, the scale $s$ and offset $o$ are given by

\begin{equation}
\label{eq_scale}
s = min\left(\frac{t_{w|h}}{p_{w|h}}, 1\right)
\end{equation}

\begin{equation}
\label{eq_scale}
o_{x|y} =  min\left(c_{w|h}, max\left(0, p_{x|y} - s \cdot p_{x|y}\right)\right)
\end{equation}

Just as the original virtual reflection module introduced by Kr\"{u}ger et al.~\cite{krueger2025virtualreflectionsdynamic2d}, the described technique has the added benefit of mapping potentially large camera images to relatively small eye models without loss of resolution. The added dynamic rescaling enables recognizable reflections of objects of arbitrary size or proximity. We implemented the mirror module using the \textit{Python} programming language~\cite{python} with the \textit{OpenCV} library~\cite{opencv_library} for image processing.

\subsubsection{Eye Expressions}

Virtual eye models enable a wide range of means for information encoding. Figure~\ref{fig:eye_expressions} shows a selection of features we implemented for our robot system. While some of these features, such as changes in pupil size or a simplified depiction of a closed eye, have counterparts in human eyes, others aim to explore expressions beyond the human repertoire. For instance, color changes may encode changes in specific system states such as turning green (Figure~\ref{fig:eye_expressions}.7) when listening to the person towards whom the eyes are directed. A subsequent appearance and disappearance of stylized iris-bands (Figure~\ref{fig:eye_expressions}.6) mimics aspects of a loading animation that can be used to indicate periods of information processing. In the \mirroreyes~condition (Figure~\ref{fig:eye_expressions}.8-10), various filters, e.g., for blur, opacity, and brightness, may be used to contextually adapt saliency or visibility of mirrored content as needed. Within the system used in the presented study some of these features were enabled. The robot used the ``smile''-expression of Figure~\ref{fig:eye_expressions}.3 after successful task completion. The loading animation was used during periods of information processing such as after the user request and upon system interruption. In the \mirroreyes~condition the brief overexposure effect (Figure~\ref{fig:eye_expressions}.10) was used when the robot first scanned objects in the scene. 

\begin{figure}[t!]
    \centering
    \includegraphics[width=1\linewidth]{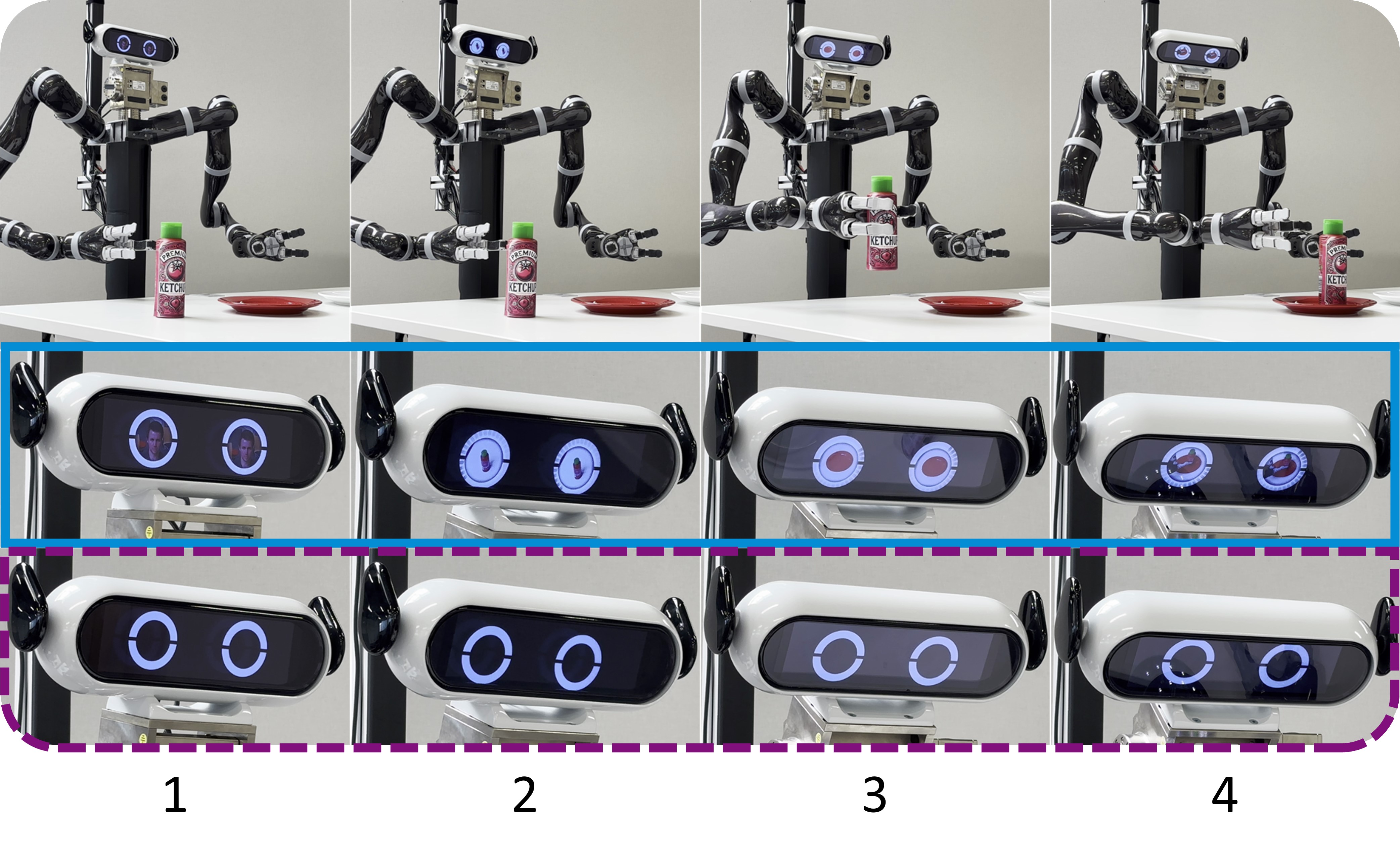}
    \caption{Exemplary action sequence and close up images of the robot head in the \mirroreyesc~(solid blue frame) and \eyesonlyc~(dashed purple frame) conditions. Sequence description: 1. The robot looks at a person standing in front of it. The person gives the instruction ``put the red bottle onto the red plate'' 2. The robot looks at the bottle it intends to pick up. 3. The robot looks at the plate it intends to place the bottle on while moving the bottle. 4. The robot looks at the joint location of the bottle and the plate after completing the pick-and-place action.}
    \label{fig:ActionSequence}
    \vspace{-10pt}
\end{figure}

\begin{table}[b!]
\caption{Action instructions that could be given to the robot}
\label{tab:requests}
\begin{adjustbox}{width=\columnwidth,center}
    \centering
    \begin{tabular}{l|l|l|l}
Instruction - \textit{Put the ...}&Expectation&Outcome&Error\\\hline
\textit{red bottle onto the red plate}&bottle~\textrightarrow~red&bottle~\textrightarrow~red&-\\
\textit{purple can onto the red plate}&can~\textrightarrow~~red&\underline{bottle}~\textrightarrow~red&1\\
\textit{spray can onto the white plate}&can~\textrightarrow~white&can~\textrightarrow~\underline{red}&2\\
\textit{spray can onto the red plate}&can~\textrightarrow~red&can~\textrightarrow~red&-\\
\textit{ketchup bottle onto the red plate}&bottle~\textrightarrow~red&\underline{can}~\textrightarrow~red&1\\  
\textit{red bottle onto the white plate}&bottle~\textrightarrow~white&bottle~\textrightarrow~\underline{red}&2\\ \hline
\end{tabular}
\end{adjustbox}
\begin{tablenotes}
\item\footnotesize Instructed actions (Expectation) and planned robot actions for the respective instruction (Outcome) are listed in a shorthand notation of the form ``object to move~\textrightarrow~color of destination plate''. \underline{Action errors are underlined}. The error column shows the classification of the action error as no error (-), a step 1 (1) or a step 2 error (2).
\end{tablenotes}
\end{table}

\subsection{User Study}

\begin{figure}[t]
    \centering
    \includegraphics[width=0.9\linewidth]{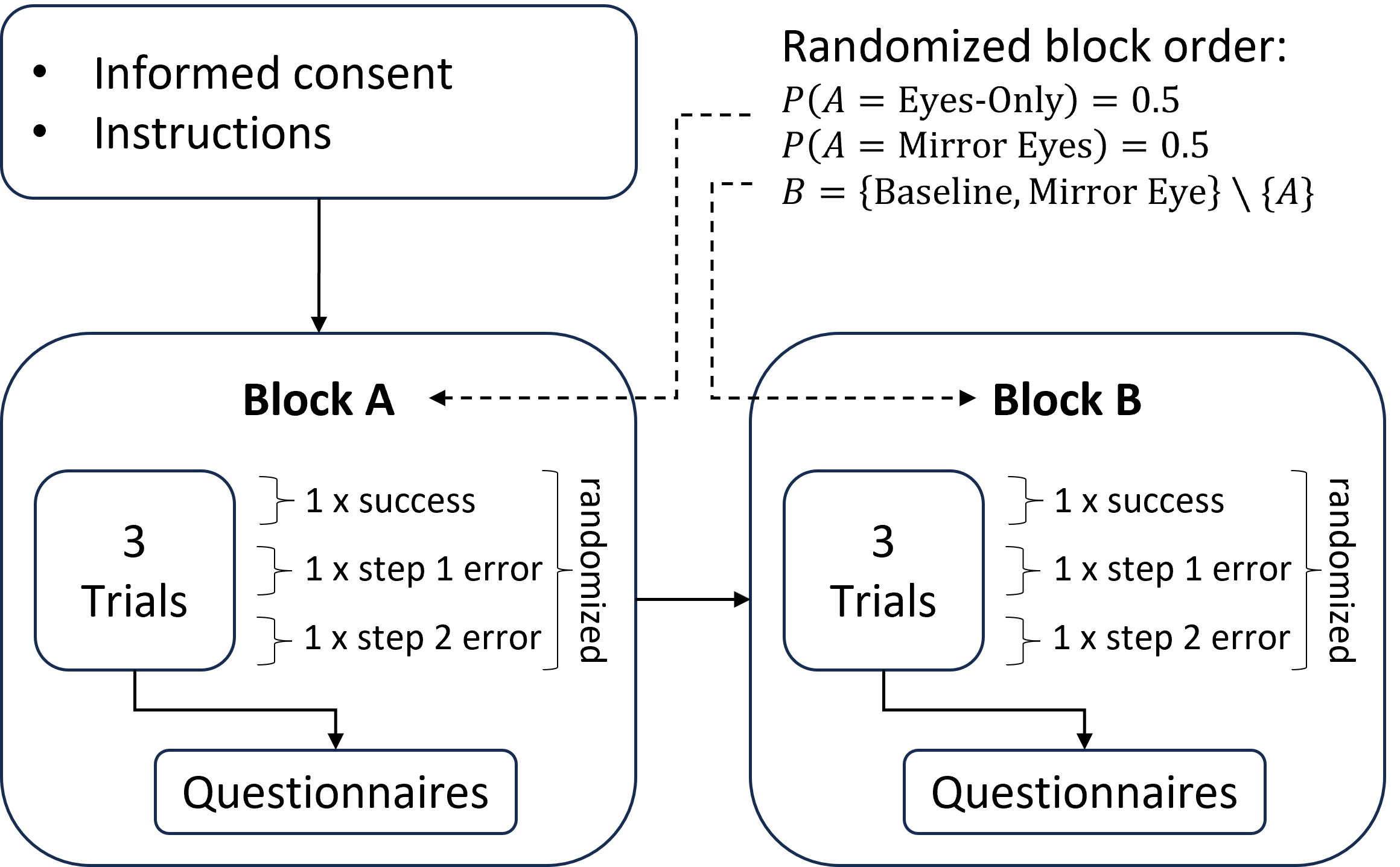}
    \caption{Experiment procedure for individual participants.}
    \label{fig:study_procedure}
    \vspace{-5pt}
\end{figure}
To investigate our hypotheses using the described robot system, we designed a human-robot interaction experiment that consisted of two blocks, which differed with respect to the configuration of the robot head. In one block, the head of the robot described in Section~\ref{sec:system} was configured to look at regions of interest through coordinated head- and eye movements (\eyesonlyc~condition). In the other block, the eyes were additionally augmented with reflection-like overlays showing mirror images of the attended regions (\mirroreyesc~condition). The difference in head appearance for both conditions can be seen in Figure \ref{fig:ActionSequence}. In both conditions, the radius of the iris was set to 1.4~cm and the radius of the pupil to 1.23~cm.

\begin{figure}[b]
    \centering
    \includegraphics[width=1\linewidth]{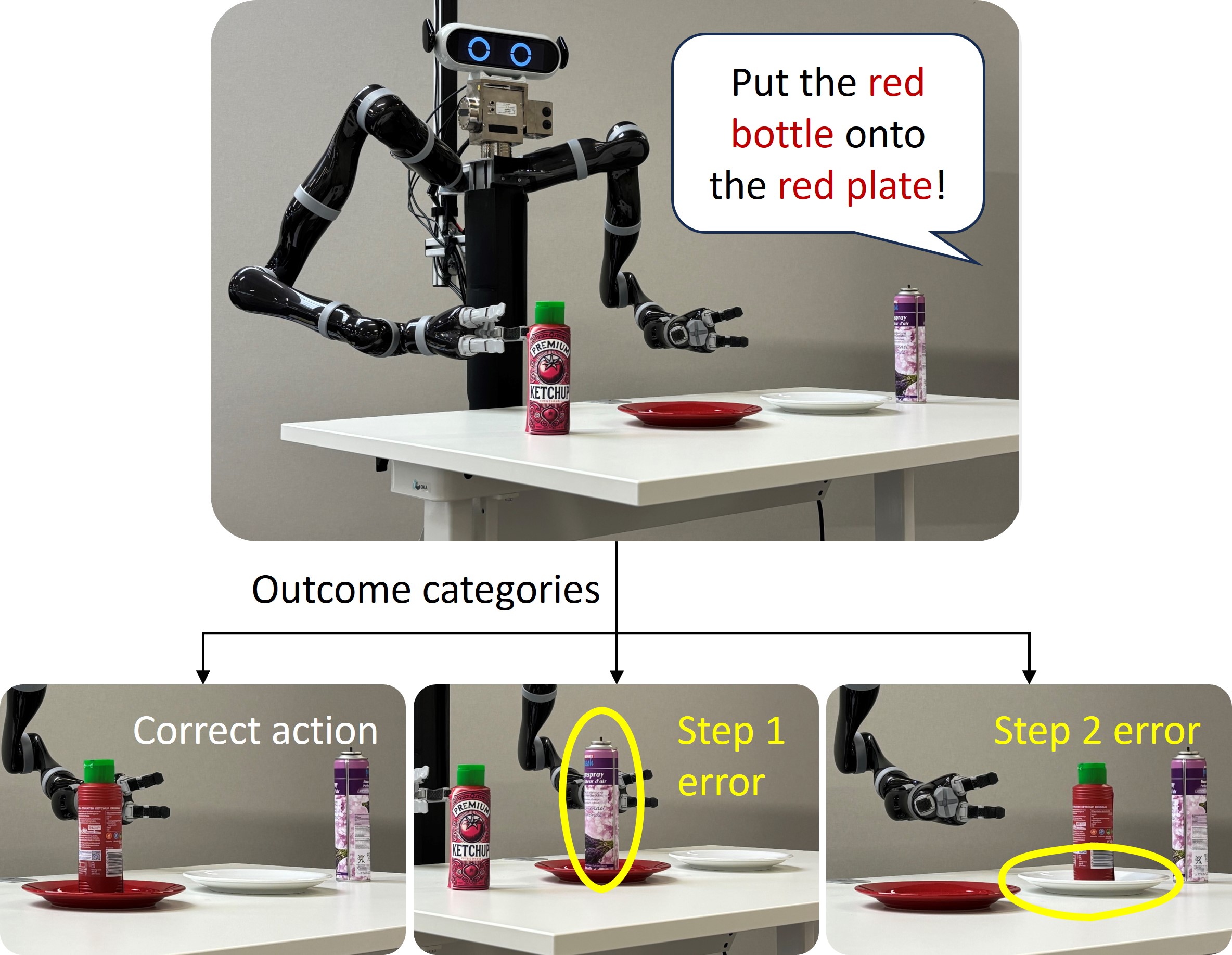}
    \caption{Potential outcomes for a specific user request without action interruption. The robot was configured to either carry out a task (speech bubble) correctly as requested (left), make an early mistake by misinterpreting the object it should pick up (middle) or make a late mistake by misinterpreting the target location of the picked object (right).}
    \label{fig:outcomes}
\end{figure}

Each experiment block consisted of three trials, in which participants were asked to stand in front of a table facing the robot, verbally make specific action requests to the robot when asked to do so by the experimenter, and monitor the robot's execution of each request. Each request involved two out of four objects available on the table such that one object (either a purple spray can or a red ketchup bottle) should be placed on another (either a red or a white plate). Table~\ref{tab:requests} lists the subset of requests that were utilized. With the listed requests, the robot could either succeed in carrying out the request or it could fail intentionally in two specific ways: 1. It could pick up the wrong object and place it on the correct target plate (Step 1 error - early). 2. It could pick up the correct object but place it on the wrong target plate (Step 2 error - late). Figure~\ref{fig:outcomes} illustrates these three potential outcome categories for a specific request. 

By prompting the LLM to misinterpret specific labels on purpose, namely \textit{purple can}, \textit{ketchup bottle} and \textit{white plate}, we enforced occurrences of the two action error types in two trials per block. For each trial, participants were asked to monitor whether the robot would carry out the given instructions correctly and interrupt the robot action in case they should notice an action error by saying \textit{stop}. Correct robot actions should not be interrupted. The order of blocks and the order of trials within each block were randomized. The durations of individual robot actions were normalized so that key events occurred at the same times relative to function onset in all trials (see Figure~\ref{fig:interruption_times_all_ecdf}). To prevent bias, participants were not instructed about the existence or purpose of the eyes in either condition. 
Exposure to each condition was intentionally kept short to minimize learning effects and assess whether understanding and utilization of the information given by the eyes would arise intuitively. 
After each block, participants were given two questionnaires to sample system explainability and user experience with respect to the robot head and their task of monitoring the robot actions for errors. Figure~\ref{fig:study_procedure} illustrates the study procedure for each participant. The procedure took about 15 minutes in total. Each block, as well as the instructions and consent took about 5 minutes.
The experiment involved 33 participants (8 women) aged between 22 and 65 years, mainly consisting of students and researchers from engineering and natural sciences disciplines.

\begin{table}[b]
\centering
  \caption{Mean stop times with standard deviations and limits (N = 33)}
  \label{tab:interruption}
  \begin{tabular}{ccccc}
    \toprule
    Error step & Condition & Mean $\pm$ SD&Minimum&Maximum\\
    \midrule
    1 & \eyesonlyc & 5.52 $\pm$ 2.77 & 2.45 & 12.73 \\ 
    1 & \mirroreyesc & 4.66 $\pm$ 1.85 & 2.38 & 9.44 \\ 
    \midrule
    2 & \eyesonlyc & 15.76 $\pm$ 1.31 & 14.09 & 19.61 \\ 
    2 & \mirroreyesc & 14.58 $\pm$ 2.9 & 4.29 & 17.71 \\ 
  \bottomrule
\end{tabular}
\end{table}

\section{RESULTS}
\label{sec:results}
\subsection{Information Processing Awareness}
We were interested in the explainability that each version of the head may provide about the system's information processing and behavior. 
For this purpose, we applied the subjective information processing awareness (SIPA) scale~\cite{Schrills2024}, which was developed to assess how well participants feel enabled by a system to perceive, understand, and predict its information processing. The SIPA score is acquired from a 6-item questionnaire, which was given to participants after completion of each experimental block.

\begin{figure}[t]
    \centering
    \includegraphics[width=0.97\linewidth]{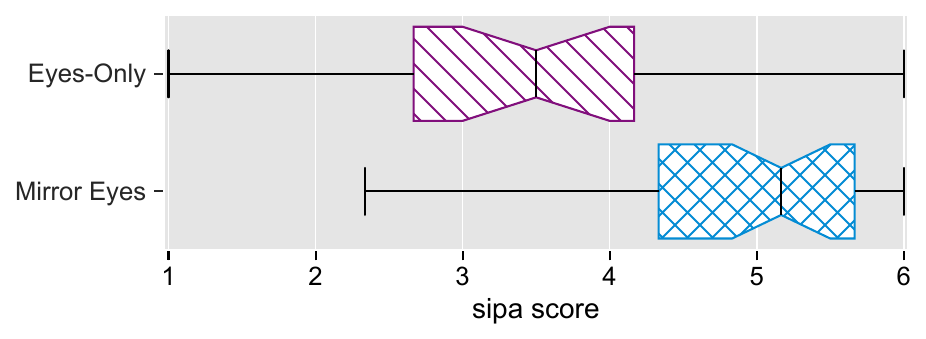}
    \caption{Subjective information processing awareness (SIPA) scores ($\Uparrow$).}
    \label{fig:sipa}
    \vspace{-10pt}
\end{figure}

Figure \ref{fig:sipa} presents the results of the SIPA ratings of 33 participants in the two display conditions of the eyes. 
SIPA scores were significantly higher ($\Uparrow$) for the \mirroreyesc~condition ($M = 4.9, SD = 1.01$) than for the \eyesonlyc~condition ($M = 3.42, SD = 1.16$), $t(32) = 8.17, p < 0.001$. In support of our first hypothesis \textbf{H1}, these results indicate that participants perceived the \mirroreyes~condition to provide additional system explainability. 

\subsection{Interruption Speed}

\begin{figure}[b]
    \centering
    \includegraphics[width=0.97\linewidth]{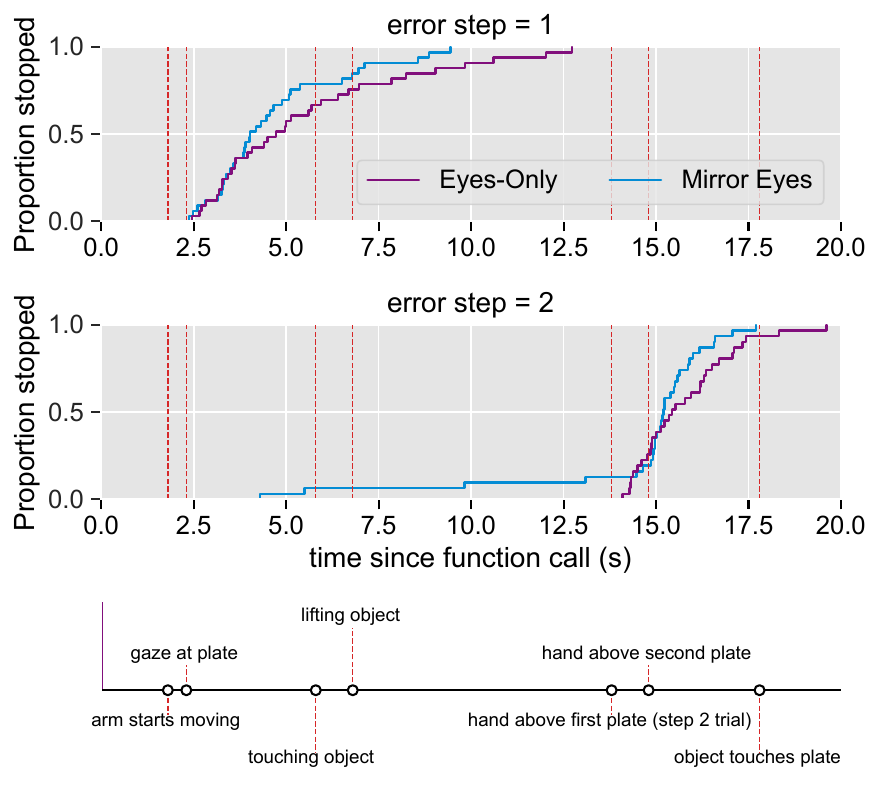}
    \caption{Cumulative distributions of error interruption times (s) between function onset and potential action completion. Dashed vertical lines indicate start times of robot actions and scene events following function onset.}
    \label{fig:interruption_times_all_ecdf}
\end{figure}

To assess people's ability to intuitively utilize the information provided by the robot eyes for constructive interference, we measured the time it took participants to interrupt the robot action when they observed the robot to make mistakes. 
The times of uttering the stop command were recorded relative to the onset of the function that would execute the respective action. 
Figure~\ref{fig:interruption_times_all_ecdf} shows the distributions of error interruption times of 33 participants across conditions from function onset until the potential action completion. Table~\ref{tab:interruption} shows the mean interruption times, standard deviations, minima, and maxima for each error step and condition. 

A two-way ANOVA revealed significant main effects for display condition ($F(1,124) = 6.1253, p<0.05$) and error step ($F(1,124)=609.6492, p<0.001$) but no significant interaction ($F(1,124)=0.1518, p=0.6975$). 
Thus, interruptions (naturally) occurred earlier for step 1 errors ($M = 5.09, SD = 2.38$) than for step 2 errors ($M = 15.17, SD = 2.31$) and participants interrupted a mistaken robot earlier in the \mirroreyesc~condition ($M = 9.47, SD = 5.54$) than in the \eyesonlyc~condition ($M = 10.32, SD = 5.7$).

These results support the hypotheses \textbf{H1} and \textbf{H2}. First, the detection of erroneous actions prior to their full execution relies on the ability to correctly interpret the robot's actions or intentions. Secondly, a mismatch between such actions or intentions and an intended goal must be recognized to warrant interference in the form of a stop command. 

\subsection{User Experience}

Having found \mirroreye-linked improvements in system explainability as well as practical advantages in terms of error detection performance, we would expect to also see a gain in the user experience.
To evaluate the user experience with the two variants of the robot head, we used the 8-item version of the User Experience Questionnaire (UEQ-S~\cite{schrepp2017design}). After each experimental block, participants filled out the UEQ-S, consisting of eight items, four representing a pragmatic scale and the other four representing a hedonic scale. Each item is rated on a 7-point semantic differential scale, ranging from -3 (most negative) to +3 (most positive).  

\begin{figure}
    \centering
    \includegraphics[width=\linewidth]{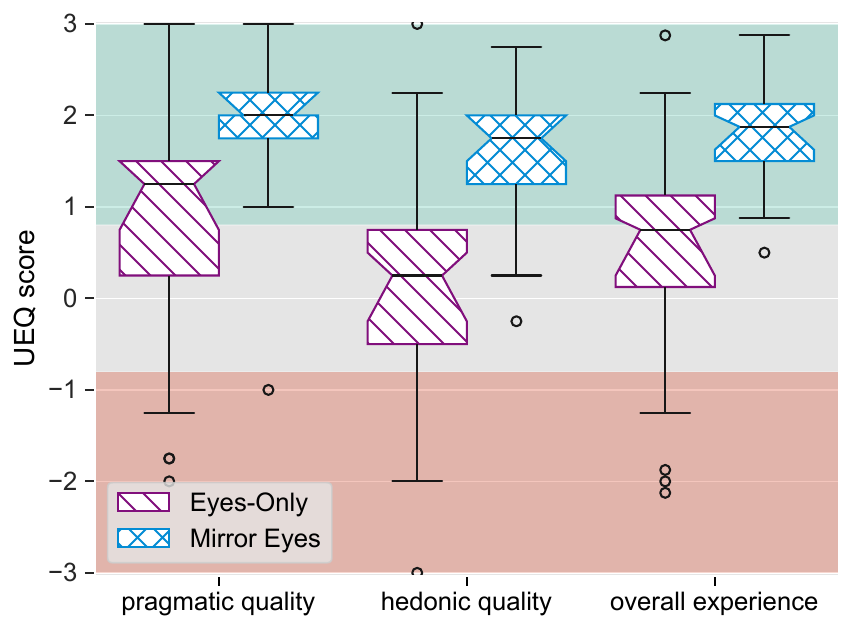}
    \caption{Box plots for UEQ-S rating of pragmatic and hedonic quality, and overall user experience across the two different conditions of the eye models. Different colors represent evaluation ranges: values between $-0.8$ and $0.8$ indicate neutral ratings (\mycolorbox{grey}{grey}), values above $0.8$ signify positive ratings (\mycolorbox{green}{green}), and values below $-0.8$ indicate negative ratings (\mycolorbox{red}{red}).}
    \label{fig:UEQ}
    \vspace{-10pt}
\end{figure}

Figure \ref{fig:UEQ} presents the results of the UEQ-S ratings of 33 participants who evaluated the user experience in the two display conditions of the eyes in terms of pragmatic quality, hedonic quality, and overall experience.  
Table \ref{tab:UEQ_score} shows the scale means and standard deviations for the UEQ-S ratings in the two display conditions of the eyes. 
Pragmatic quality scores were significantly higher for the \mirroreyesc~condition ($M=2.01$, $SD=0.72$) than for the \eyesonlyc~condition ($M=0.88$, $SD=1.32$), $t(32)=5.32$, $p < 0.001$. Also, hedonic quality scores were significantly higher for the \mirroreyesc~condition ($M = 1.67$, $SD = 0.66$) than for the \eyesonlyc~condition ($M=0.08, SD=1.32$), $t(32)=8.11$, $p < 0.001$. Consequently, the overall user experience was significantly higher for the \mirroreyesc~condition ($M=1.84$, $SD=0.52$) than for the \eyesonlyc~condition ($M=0.48$, $SD=0.52$), $t(32)=8.22$, $p < 0.001$. 

In relation to a benchmark dataset of 468 studies~\cite{hinderks2018benchmark}, these ratings put the user experience in the \mirroreyes~condition into the ``Excellent'' category and within the top 10\% range of all studies in the user experience benchmark dataset. These results support hypothesis \textbf{H3} about the positive impact of \mirroreyes~on user experience in human-robot interaction. 

\begin{table}[t]
\centering
  \caption{Score means and standard deviations of the UEQ-S (N = 33)}
  \label{tab:UEQ_score}
  \begin{tabular}{lccc}
    \toprule
    Condition & Pragmatic quality & Hedonic quality & Overall\\
    \midrule
    \eyesonlyc & \positive{0.88 $\pm$ 1.32} & \neutral{0.08 $\pm$ 1.32} & \neutral{0.48 $\pm$ 1.17} \\ 
    \mirroreyesc & \positive{2.01 $\pm$ 0.72} & \positive{1.67 $\pm$ 0.66} & \positive{1.84 $\pm$ 0.52} \\
  \bottomrule
\end{tabular}
\vspace{-10pt}
\end{table}

\section{DISCUSSION}

Here we introduced a system that embeds the \mirroreyes~concept~\cite{krueger2025virtualreflectionsdynamic2d} in a robot equipped with a moving head, speech processing, vision processing, multimodal reasoning, and physical manipulation capabilities to investigate potential effects of \mirroreyes~on human-robot interaction settings. 
We conducted a human-robot interaction study with 33 participants in which participants were asked to verbally instruct the robot to carry out specific pick-and-place tasks, monitor the robot's execution of the request for correctness, and interrupt it verbally in case of erroneous actions. The task was carried out in two experimental conditions, one with an enabled \mirroreye~feature and one that included the eyes but lacked the mirroring effect. In one third of the trials the robot was programmed to make an error early during the pick-up phase, in one third it made a late error during the object placement phase, and in the last third it carried out the request correctly. We found that participants felt more aware about the robot's information processing, detected erroneous actions earlier, and rated the user experience higher when mirroring was enabled. These results are consistent with our hypothesis about a transfer of \mirroreye~benefits to physically embedded robot systems and confirm that the benefits are not exclusive to faces as reference targets. In fact, compared to findings on screen-based \mirroreye~use in a group interaction setting~\cite{krueger2025virtualreflectionsdynamic2d}, the user experience with the \mirroreyes~embedded in a physically moving robot head was even higher in the present study. 

One possible explanation for the transfer and even amplification of \mirroreye~benefits lies in the nature of human multimodal perception. Human sensory experience is partly attributable to an integration of information from multiple senses, including percepts that might feel strictly unimodal~\cite{McGurk1976,Botvinick1998,Verhagen2006} (see ~\cite{Krueger2022}). 
Even in the absence of ambiguity, congruent~\cite{Meredith1987,Stein1993,Meredith1996,Alais2010} stimuli from multiple senses have been found to be advantageous, e.g., by increasing reaction speed~\cite{Todd1912,Hershenson1962,Nickerson1973,Bernstein1969,Diederich2004,Mozolic2012} and reducing the risk of sensory overload in demanding scenarios~\cite{Hancock2013, Wahn2016}.
A virtual reflection and an eye model are both perceived visually. Nevertheless, physiological models for visual processing suggest that they might not be processed in the same way:  
The two-stream model for visual processing distinguishes between two main neural pathways that process different aspects of visual information. The ventral stream processes foveal retinal input with a high spatial resolution for the recognition of objects and faces~\cite{Kanwisher2002fusiform} whereas the dorsal stream is responsible for processing motion and spatial relationships, including observed eye movements~\cite{braddick2003normal} at high temporal resolution using input from the entire retina. These complementary and parallel roles might be exploited by stimuli that combine location and identity information, such as those created by the \mirroreyes. But even if such a within-sense facilitation should not be present or relevant, \mirroreyes~could still be supportive for a disambiguation of information from other senses such as speech and other sounds. 

One key factor to take into account in the interpretation of the results is that they reflect interactions with very brief exposure to the system and each experimental condition. This restriction, coupled with the intentional omission of any explanation about the functionality or role of the eyes, was implemented to capture the intuition the \mirroreyes~might provide in human-robot interaction. The results suggest, that an intuitive use of the \mirroreyes~has indeed taken place and was met with high appreciation by the participants. 

Nevertheless, this avoidance of learning effects also means that effects might change once participants get used to the presence of the eyes in either condition and learn what features to pay attention to at what points during an interaction. 
A first glance into the potential of more optimized use by human interaction partners might be present in the interruption times of a subset of participants in step 2 (late) error cases. Figure~\ref{fig:interruption_times_all_ecdf} shows that some participants were able to identify and interrupt step 2 errors around 10 seconds earlier in the \mirroreyes~condition than in the \eyesonly~condition. This was theoretically possible because the gaze on the placement destination preceded the corresponding movement. We observed that most participants in the \mirroreyes~condition appeared to have noticed the gaze towards the wrong placement target in step 2 error cases early, but hesitated with the stop command until receiving confirmation from a corresponding movement of the arms. Similarly, for pick-up errors (step 1), participants who showed signs of early error detection through the \mirroreyes~typically appeared to wait for arm movement confirmation before interruption. 
We therefore consider it likely that the proportion of early interruptions will increase once participants become more familiar with the scenario, e.g., by increasing the number of trials per participant. In contrast to the present study, which focused on early intuition, future research should attend to the development of \mirroreye~effects after prolonged or repeated exposure, as well as the role of robot kinematics in disambiguating eye behavior. 

One reason for hesitation in acting earlier on information provided by the \mirroreyes~could be the multitude of roles eyes play as communication interfaces: Eyes point towards regions that are observed, but the reasons for observation can be diverse, such as scanning the environment or preparing an action. While the \mirroreyes~may leave little doubt about \textit{what} is observed, there can still be ambiguity about \textit{why} it is observed. To an extent, we tried to counter this ambiguity by introducing brief moments of overexposure during initial object registration, as depicted in Figure \ref{fig:eye_expressions}.10. However, since no element of the eyes was explained to the participants, it is unclear whether the participants noticed or interpreted this detail in fulfillment of their monitoring task. 
Future work could focus on extensions that aim to circumvent remaining ambiguities while preserving the observed intuition. 

\section{CONCLUSION}

The addition of a reflection-like overlay on the display of virtual eye models (\mirroreyes) on a mobile robot head significantly improved people's subjective understanding of a robot's information processing, their ability to detect and interrupt erroneous robot actions, and their user experience with the robot head as a communication channel in interaction with the robot. These results suggest that even mobile robot heads, which can be physically oriented towards regions of interest, benefit from \mirroreye~utilization in human-robot interaction scenarios. Because these effects occurred after very short exposure to the system and without any explanations about the role of the eyes in the interaction, we believe that they are indicative of the presence of quick intuition. Future research should investigate the effects of prolonged exposure to \mirroreyes~as well as complementary means for reducing any remaining ambiguities the eyes may have as explanatory media. 

\addtolength{\textheight}{-0cm}   




\bibliographystyle{IEEEtran}
\bibliography{mirror_eye_lit}

\begin{thebibliography}{10}
\providecommand{\url}[1]{#1}
\csname url@samestyle\endcsname
\providecommand{\newblock}{\relax}
\providecommand{\bibinfo}[2]{#2}
\providecommand{\BIBentrySTDinterwordspacing}{\spaceskip=0pt\relax}
\providecommand{\BIBentryALTinterwordstretchfactor}{4}
\providecommand{\BIBentryALTinterwordspacing}{\spaceskip=\fontdimen2\font plus
\BIBentryALTinterwordstretchfactor\fontdimen3\font minus \fontdimen4\font\relax}
\providecommand{\BIBforeignlanguage}[2]{{%
\expandafter\ifx\csname l@#1\endcsname\relax
\typeout{** WARNING: IEEEtran.bst: No hyphenation pattern has been}%
\typeout{** loaded for the language `#1'. Using the pattern for}%
\typeout{** the default language instead.}%
\else
\language=\csname l@#1\endcsname
\fi
#2}}
\providecommand{\BIBdecl}{\relax}
\BIBdecl

\bibitem{Bratman1992shared}
M.~E. Bratman, ``Shared cooperative activity,'' \emph{The philosophical review}, 1992.

\bibitem{Hoc2001towards}
J.-M. Hoc, ``Towards a cognitive approach to human--machine cooperation in dynamic situations,'' \emph{International journal of human-computer studies}, 2001.

\bibitem{Krueger2017}
M.~Kr{\"u}ger, C.~B. Wiebel, and H.~Wersing, ``From tools towards cooperative assistants,'' in \emph{International Conference on Human Agent Interaction}, 2017.

\bibitem{Sendhoff2020}
B.~Sendhoff and H.~Wersing, ``Cooperative intelligence-a humane perspective,'' in \emph{IEEE international conference on human-machine systems (ICHMS)}, 2020.

\bibitem{Wollstadt2022quantifying}
P.~Wollstadt and M.~Kr{\"u}ger, ``Quantifying cooperation between artificial agents using synergistic information,'' in \emph{IEEE Symposium Series on Computational Intelligence (SSCI)}, 2022.

\bibitem{vaswani2017attention}
A.~Vaswani, N.~Shazeer, N.~Parmar, J.~Uszkoreit, L.~Jones, A.~N. Gomez, {\L}.~Kaiser, and I.~Polosukhin, ``Attention is all you need,'' \emph{Neural information processing systems}, 2017.

\bibitem{brown2020language}
T.~Brown, B.~Mann, N.~Ryder, M.~Subbiah \emph{et~al.}, ``Language models are few-shot learners,'' \emph{Neural information processing systems}, 2020.

\bibitem{Wang2024Lami}
C.~Wang, S.~Hasler, D.~Tanneberg, F.~Ocker, F.~Joublin, A.~Ceravola, J.~Deigmoeller, and M.~Gienger, ``Lami: Large language models for multi-modal human-robot interaction,'' in \emph{Extended Abstracts of the CHI Conference on Human Factors in Computing Systems}, 2024.

\bibitem{tanneberg2024help}
D.~Tanneberg, F.~Ocker, S.~Hasler, J.~Deigmoeller, A.~Belardinelli, C.~Wang, H.~Wersing, B.~Sendhoff, and M.~Gienger, ``To help or not to help: Llm-based attentive support for human-robot group interactions,'' in \emph{IEEE/RSJ International Conference on Intelligent Robots and Systems (IROS)}, 2024.

\bibitem{Stiefelhagen2004}
R.~Stiefelhagen, C.~Fugen, R.~Gieselmann, H.~Holzapfel, K.~Nickel, and A.~Waibel, ``Natural human-robot interaction using speech, head pose and gestures,'' in \emph{IEEE/RSJ International Conference on Intelligent Robots and Systems (IROS)}, 2004.

\bibitem{Maurtua2017multimodal}
I.~Maurtua, I.~Fernandez, A.~Tellaeche, J.~Kildal, L.~Susperregi, A.~Ibarguren, and B.~Sierra, ``Natural multimodal communication for human--robot collaboration,'' \emph{International Journal of Advanced Robotic Systems}, 2017.

\bibitem{Wang2020watchout}
C.~Wang, M.~Kr{\"u}ger, and C.~B. Wiebel-Herboth, ``“watch out!”: prediction-level intervention for automated driving,'' in \emph{International Conference on Automotive User Interfaces and Interactive Vehicular Applications}, 2020.

\bibitem{menendez2025semanticscanpathcombininggazespeech}
E.~Menendez, M.~Gienger, S.~Martínez, C.~Balaguer, and A.~Belardinelli, ``Semanticscanpath: Combining gaze and speech for situated human-robot interaction using llms,'' 2025.

\bibitem{belardinelli2025}
A.~Belardinelli, C.~Wang, D.~Tanneberg, M.~Kr\"uger, S.~Hasler, and M.~Gienger, ``Explaining to and being explained by a service robot: Four hri studies revisited under a framework for explainability,'' in \emph{Proceedings of the 3rd TRR Conference Contextualizing Explanations (ContEx)}, 2025.

\bibitem{Salem2012generation}
M.~Salem, S.~Kopp, I.~Wachsmuth, K.~Rohlfing, and F.~Joublin, ``Generation and evaluation of communicative robot gesture,'' \emph{International Journal of Social Robotics}, 2012.

\bibitem{Quintero2015}
C.~P. Quintero, R.~Tatsambon, M.~Gridseth, and M.~Jägersand, ``Visual pointing gestures for bi-directional human robot interaction in a pick-and-place task,'' in \emph{IEEE International Symposium on Robot and Human Interactive Communication (RO-MAN)}, 2015.

\bibitem{Hoffman2015design}
G.~Hoffman, O.~Zuckerman, G.~Hirschberger, M.~Luria, and T.~Shani~Sherman, ``Design and evaluation of a peripheral robotic conversation companion,'' in \emph{ACM/IEEE international conference on human-robot interaction}, 2015.

\bibitem{Ishi2018}
C.~T. Ishi, D.~Machiyashiki, R.~Mikata, and H.~Ishiguro, ``A speech-driven hand gesture generation method and evaluation in android robots,'' \emph{IEEE Robotics and Automation Letters}, 2018.

\bibitem{Metta2008}
G.~Metta, G.~Sandini, D.~Vernon, L.~Natale, and F.~Nori, ``The icub humanoid robot: An open platform for research in embodied cognition,'' in \emph{8th Workshop on Performance Metrics for Intelligent Systems}, 2008.

\bibitem{Zaraki2014gazecontrol}
A.~Zaraki, D.~Mazzei, M.~Giuliani, and D.~De~Rossi, ``Designing and evaluating a social gaze-control system for a humanoid robot,'' \emph{IEEE Transactions on Human-Machine Systems}, 2014.

\bibitem{admoni2017social}
H.~Admoni and B.~Scassellati, ``Social eye gaze in human-robot interaction: a review,'' \emph{Journal of Human-Robot Interaction}, 2017.

\bibitem{Yoshida2022}
N.~Yoshida, S.~Yonemura, M.~Emoto, K.~Kawai, N.~Numaguchi, H.~Nakazato, S.~Otsubo, M.~Takada, and K.~Hayashi, ``Production of character animation in a home robot: A case study of lovot,'' \emph{International Journal of Social Robotics}, 2022.

\bibitem{belardinelli2022intention}
A.~Belardinelli, A.~R. Kondapally, D.~Ruiken, D.~Tanneberg, and T.~Watabe, ``Intention estimation from gaze and motion features for human-robot shared-control object manipulation,'' in \emph{IEEE/RSJ International Conference on Intelligent Robots and Systems (IROS)}, 2022.

\bibitem{krueger2025virtualreflectionsdynamic2d}
M.~Kr{\"u}ger, Y.~Oshima, and Y.~Fang, ``Virtual reflections on a dynamic 2d eye model improve spatial reference identification,'' \emph{arXiv}, 2025.

\bibitem{teyssier2021eyecam}
M.~Teyssier, M.~Koelle, P.~Strohmeier, B.~Fruchard, and J.~Steimle, ``Eyecam: Revealing relations between humans and sensing devices through an anthropomorphic webcam,'' in \emph{CHI Conference on Human Factors in Computing Systems}, 2021.

\bibitem{Gomez2018}
R.~Gomez, D.~Szapiro, K.~Galindo, and K.~Nakamura, ``Haru: Hardware design of an experimental tabletop robot assistant,'' in \emph{ACM/IEEE International Conference on Human-Robot Interaction}, 2018.

\bibitem{fang2023designing}
Y.~Fang, L.~Merino, S.~Thill, and R.~Gomez, ``Designing visual and auditory attention-driven movements of a tabletop robot,'' in \emph{IEEE International Conference on Robot and Human Interactive Communication (RO-MAN)}, 2023.

\bibitem{Fang2024Roman}
Y.~Fang, J.~M. Pérez-Molerón, L.~Merino, and R.~Gomez, ``Enhancing human perception of direct gaze from a social robot through eye-head coordination,'' in \emph{IEEE International Conference on Robot and Human Interactive Communication (RO-MAN)}, 2024.

\bibitem{Mccarthy1997}
G.~McCarthy, A.~Puce, J.~C. Gore, and T.~Allison, ``Face-specific processing in the human fusiform gyrus,'' \emph{Journal of cognitive neuroscience}, 1997.

\bibitem{besson2017}
G.~Besson, G.~Barragan-Jason, S.~J. Thorpe, M.~Fabre-Thorpe, S.~Puma, M.~Ceccaldi, and E.~J. Barbeau, ``From face processing to face recognition: Comparing three different processing levels,'' \emph{Cognition}, 2017.

\bibitem{George2001}
N.~George, J.~Driver, and R.~J. Dolan, ``Seen gaze-direction modulates fusiform activity and its coupling with other brain areas during face processing,'' \emph{Neuroimage}, 2001.

\bibitem{joublin2024copal}
F.~Joublin, A.~Ceravola, P.~Smirnov, F.~Ocker, J.~Deigmoeller, A.~Belardinelli, C.~Wang, S.~Hasler, D.~Tanneberg, and M.~Gienger, ``{CoPAL}: corrective planning of robot actions with large language models,'' in \emph{IEEE International Conference on Robotics and Automation (ICRA)}, 2024.

\bibitem{ravi2024sam2}
N.~Ravi, V.~Gabeur, Y.-T. Hu, R.~Hu, C.~Ryali, T.~Ma, H.~Khedr, R.~R{\"a}dle, C.~Rolland, L.~Gustafson, E.~Mintun, J.~Pan, K.~V. Alwala, N.~Carion, C.-Y. Wu, R.~Girshick, P.~Doll{\'a}r, and C.~Feichtenhofer, ``Sam 2: Segment anything in images and videos,'' \emph{arXiv}, 2024.

\bibitem{nakamura1991advanced}
Y.~Nakamura, \emph{Advanced Robotics: Redundancy and Optimization}.\hskip 1em plus 0.5em minus 0.4em\relax Addison-Wesley, 1991.

\bibitem{shibata2001biomimetic}
T.~Shibata, S.~Vijayakumar, J.~Conradt, and S.~Schaal, ``Biomimetic oculomotor control,'' \emph{Adaptive Behavior}, 2001.

\bibitem{python}
G.~Van~Rossum and F.~L. Drake~Jr, \emph{Python reference manual}.\hskip 1em plus 0.5em minus 0.4em\relax Centrum voor Wiskunde en Informatica Amsterdam, 1995.

\bibitem{opencv_library}
G.~Bradski, ``{The OpenCV Library},'' \emph{Dr. Dobb's Journal of Software Tools}, 2000.

\bibitem{Schrills2024}
T.~Schrills, M.~Sieger, M.~Gruner, and T.~Franke, ``Evaluation of a scale to assess subjective information processing awareness of humans in interaction with automation \& artificial intelligence,'' \emph{Artificial Intelligence and Social Computing}, 2024.

\bibitem{schrepp2017design}
M.~Schrepp, J.~Thomaschewski, and A.~Hinderks, ``Design and evaluation of a short version of the user experience questionnaire (ueq-s),'' \emph{International Journal of Interactive Multimedia and Artificial Intelligence}, 2017.

\bibitem{hinderks2018benchmark}
A.~Hinderks, M.~Schrepp, and J.~Thomaschewski, ``A benchmark for the short version of the user experience questionnaire,'' in \emph{International Conference on Web Information Systems and Technologies-APMDWE}, 2018.

\bibitem{McGurk1976}
H.~McGurk and J.~MacDonald, ``Hearing lips and seeing voices,'' \emph{Nature}, 1976.

\bibitem{Botvinick1998}
M.~Botvinick, J.~Cohen \emph{et~al.}, ``Rubber hands ' feel' touch that eyes see,'' \emph{Nature}, 1998.

\bibitem{Verhagen2006}
J.~V. Verhagen and L.~Engelen, ``The neurocognitive bases of human multimodal food perception: sensory integration,'' \emph{Neuroscience \& biobehavioral reviews}, 2006.

\bibitem{Krueger2022}
M.~Kr{\"u}ger, ``An enactive approach to perceptual augmentation in mobility,'' Ph.D. dissertation, Ludwig-Maximilians-Universität München, 2022.

\bibitem{Meredith1987}
M.~Meredith, J.~Nemitz, and B.~Stein, ``Determinants of multisensory integration in superior colliculus neurons. i. temporal factors,'' \emph{Journal of Neuroscience}, 1987.

\bibitem{Stein1993}
B.~E. Stein and M.~A. Meredith, \emph{The merging of the senses.}\hskip 1em plus 0.5em minus 0.4em\relax The MIT Press, 1993.

\bibitem{Meredith1996}
M.~A. Meredith and B.~E. Stein, ``Spatial determinants of multisensory integration in cat superior colliculus neurons,'' \emph{Journal of Neurophysiology}, 1996.

\bibitem{Alais2010}
D.~Alais, F.~Newell, and P.~Mamassian, ``Multisensory processing in review: from physiology to behaviour,'' \emph{Seeing and Perceiving}, 2010.

\bibitem{Todd1912}
J.~W. Todd, \emph{Reaction to multiple stimuli}.\hskip 1em plus 0.5em minus 0.4em\relax Science Press, 1912.

\bibitem{Hershenson1962}
M.~Hershenson, ``Reaction time as a measure of intersensory facilitation.'' \emph{Journal of experimental psychology}, 1962.

\bibitem{Nickerson1973}
R.~S. Nickerson, ``Intersensory facilitation of reaction time: energy summation or preparation enhancement?'' \emph{Psychological review}, 1973.

\bibitem{Bernstein1969}
I.~H. Bernstein, M.~H. Clark, and B.~A. Edelstein, ``Effects of an auditory signal on visual reaction time.'' \emph{Journal of experimental psychology}, 1969.

\bibitem{Diederich2004}
A.~Diederich and H.~Colonius, ``Bimodal and trimodal multisensory enhancement: Effects of stimulus onset and intensity on reaction time,'' \emph{Perception {\&} Psychophysics}, 2004.

\bibitem{Mozolic2012}
J.~L. Mozolic, C.~E. Hugenschmidt, A.~M. Peiffer, and P.~J. Laurienti, ``Multisensory integration and aging,'' \emph{In: The Neural Bases of Multisensory Processes}, 2012.

\bibitem{Hancock2013}
P.~A. Hancock, J.~E. Mercado, J.~Merlo, and J.~B.~V. Erp, ``Improving target detection in visual search through the augmenting multi-sensory cues,'' \emph{Ergonomics}, 2013.

\bibitem{Wahn2016}
B.~Wahn, J.~Schwandt, M.~Kr\"uger, D.~Crafa, V.~Nunnendorf, and P.~K\"onig, ``Multisensory teamwork: using a tactile or an auditory display to exchange gaze information improves performance in joint visual search,'' \emph{Ergonomics}, 2016.

\bibitem{Kanwisher2002fusiform}
N.~Kanwisher, J.~McDermott, and M.~M. Chun, ``The fusiform face area: a module in human extrastriate cortex specialized for face perception,'' \emph{Journal of neuroscience}, 1997.

\bibitem{braddick2003normal}
O.~Braddick, J.~Atkinson, and J.~Wattam-Bell, ``Normal and anomalous development of visual motion processing: motion coherence and ‘dorsal-stream vulnerability’,'' \emph{Neuropsychologia}, 2003.

\end{thebibliography}

\end{document}